# The quantization error in a Self-Organizing Map as a contrast and color specific indicator of single-pixel change in large random patterns

John M. Wandeto[+] and Birgitta Dresp-Langley[*]

[*]Centre National de la Recherche Scientifique (CNRS), UMR 7357 ICube Lab, University of Strasbourg, FRANCE

[+]Dedan Kimathi University of Technology, Department of Information Technology, Nyeri, KENYA

*Corresponding author: birgitta.dresp@unistra.fr

**Abstract**

The quantization error in a fixed-size Self-Organizing Map (SOM) with unsupervised winner-take-all learning has previously been used successfully to detect, in minimal computation time, highly meaningful changes across images in medical time series and in time series of satellite images. Here, the functional properties of the quantization error in SOM are explored further to show that the metric is capable of reliably discriminating between the finest differences in local contrast intensities and contrast signs. While this capability of the QE is akin to functional characteristics of a specific class of retinal ganglion cells (the so-called Y-cells) in the visual systems of the primate and the cat, the sensitivity of the QE surpasses the capacity limits of human visual detection. Here, the quantization error in the SOM is found to reliably signal changes in contrast or colour when contrast information is removed from or added to the image, but not when the amount and relative weight of contrast information is constant and only the local spatial position of contrast elements in the pattern changes. While the RGB Mean reflects coarser changes in colour or contrast well enough, the SOM-QE is shown to outperform the RGB Mean in the detection of single-pixel changes in images with up to five million pixels. This could have important implications in the context of unsupervised image learning and computational building block approaches to large sets of image data (*big data*), including deep learning blocks, and automatic detection of contrast change at the nanoscale in Transmission or Scanning Electron Micrographs (TEM, SEM), or at the subpixel level in multispectral and hyper-spectral imaging data.

**Introduction**

Sensitivity to the intensity, spatial extent, and polarity of physical contrast are two major functional characteristics of natural visual systems, found in primates and other superior mammals (Hubel, 1963; Hubel and Wiesel, 1959, 1965, 1968). A large amount of Steve Grossberg's theoretical work has been devoted to developing empirically inspired neural models for visual analysis and perception-based image interpretation using these, fundamentally important, functional properties of biological visual neurons (Grossberg, 1997). His work with close colleagues and collaborators most successfully implemented them in increasingly complex and powerful neural network architectures to predict how the human brain extracts meaning, in terms of 3D objects and their different specific qualities, from a seemingly random variability of contrast information in 2D images (e.g. Grossberg, 2013, 2015; Grossberg, Srinivasan, and Yazdabakhsh, 2011; Grossberg and Pinna, 2012, and many others, before and after). A large amount of literature from the behavioural neurosciences has confirmed that the same functional properties as those explored in Steve Grossberg's model approaches ensure the visual perceptual processing of meaningful 2D image contents, thereby enabling their further cognitive analysis (e.g. Zhang and von der Heydt, 2010; Su, He, and Ooi, 2010; Spillmann, Dresp-Langley, and Tseng, 2015; Dresp-Langley and Grossberg, 2016, and many others, before and after).

This contribution to Steve Grossberg's Special Birthday Issue is to be seen as a celebration of his extraordinarily rich and complex work and achievement as a neural modelling expert and visual scientist. Our article here explores previously unsuspected functional properties of a neural network metric not many studies have focused on: the quantization error (QE) in the Self-Organizing Map (SOM). The QE in the SOM, or SOM-QE, is a statistical metric that represents the difference between data and results obtained by letting a Self-Organizing neural network learn the data. In its most generic form a SOM, also known as a Kohonen map (Kohonen, 1981), is a formal neural network for visual pattern analysis. In previous work by others, the quantization error from SOM output has been exploited as a qualitative neural network metric, either in comparison with vector quantization in variable neural network architectures used on the same image contents (Kohonen, Nieminen, and Honkela, 2009) or, more recently as a metric for dynamic precision scaling in neural network training (Taras and Stuart, 2018).

The Self-Organizing Map (a prototype is graphically represented here in Figure 1 for illustration) may be described formally as a nonlinear, ordered, smooth mapping of high-dimensional input data onto the elements of a regular, low-dimensional array (Kohonen, 2001). Assume that the set of input variables is definable as a real vector *x*, of n-dimension. With each element in the SOM array, we associate a parametric real vector $m_i$, of n-dimension. $m_i$ is called a model, hence the SOM array is composed of models. Assuming a general distance measure between *x* and $m_i$ denoted by *d(x,$m_i$)*, the map of an input vector *x* on the SOM array is defined as the array element $m_c$ that matches best (smallest *d(x,$m_i$)*) with *x*. During the learning process, the input vector *x* is compared with all the $m_i$ in order to identify its $m_c$. The Euclidean distances ||*x*-$m_i$|| define $m_c$. Models that are topographically close in the map up to a certain geometric distance, denoted by $h_{ci}$, will activate each other to learn something from the same input *x*. This will result in a local relaxation or smoothing effect on the models in this neighborhood, which in continued learning leads to global ordering. SOM learning is represented by the equation

$$m(t+1) = m_i(t) + \alpha(t)h_{ci}(t)[x(t) - m_i(t)] \qquad (1)$$

where $t = 1,2,3...$is an integer, the discrete-time coordinate, $h_{ci}(t)$ is the neighborhood function, a smoothing kernel defined over the map points which converges towards zero with time, $\alpha(t)$ is the learning rate, which also converges towards with time and affects the amount of learning in each model. At the end of the *winner-take-all* learning process in the SOM, each image input vector *x* becomes associated to its best matching model on the map $m_c$. The difference between *x* and $m_c$, ||*x*-$m_c$||, is a measure of how close the final SOM value is to the original input value and is reflected by the quantization error QE. The QE of *x* is given by

$$QE = 1/N \sum_{i=1}^{N} \|X_i - m_{c_i}\| \qquad (2)$$

where N is the number of input vectors *x* in the image. The final weights of the SOM are defined by a three dimensional output vector space representing each R, G, and B channel. The magnitude as well as the direction of change in any of these from one image to another is reliably reflected by changes in the QE.

In previous studies (Wandeto, Nyongesa, and Dresp-Langley, 2016; Dresp-Langley, Wandeto, and Nyongesa, 2018a), we used one and the same SOM architecture on series of computer generated random-dot images with systematically introduced changes across images. The neural network (SOM) was always trained on the first image (*reference*) of a given series. We discovered that the SOM-QE, generated by submitting the other images (*tests*) from the same series to the same neural network analysis, displayed 1) statistically consistent and significant sensitivity to changes in local contrast intensity when spatial extent of contrast is constant across images, 2) statistically consistent and significant sensitivity to changes in local extent of contrast when contrast intensity is constant across images. It was concluded that the SOM-QE is a statistically reliable indicator of contrast changes in image series. The SOM-QE was subsequently exploited by our group for the computational analysis of image time series in two different applied contexts.

1) In the context of medical image analysis for detecting clinically relevant local changes across image contents; the Hautepierre University Hospital Center (CHU Strasbourg, courtesy Dr. Philippe Choquet and his team) provided us with time series of MRI images from two distinct clinical examinations of the knee of one and the same patient, before and after a minor accident. The latter had produced localized tissue lesions around the bones of the knee-joint that were not immediately detected by the expert radiologist, although the patient had experienced intense pain. In a clinical context of image interpretation, the first problem is being able to decide whether meaningful image contrasts in the clinically relevant image regions are likely to have changed between examinations or not. Such change is reflected either by a local increase in spatial image contrast, indicating pathological tissue alteration due to an inflammatory process, a traumatic lesion, or a growing tumour, or by a local decrease in spatial image contrast, possibly indicating that the given pathology may be receding. By submitting the two MRI image series from the two clinical examinations of our patient to SOM analysis, we found a systematic difference in QE from analyses of the first series and analyses of the second series (Wandeto, Nyongesa, Remond, and Dresp-Langley, 2017). Between the two examinations, local image contrasts had changed consistently and significantly, indicating local tissue inflammation caused by the minor accident that had happened between the two clinical exams. These local contrast changes were reliably captured by the variations in the SOM-QE (Figure 2).

2) In the context of image-based environmental change detection; the same approach was applied to time series of satellite images of the Nevada Desert nearby Las Vegas City, USA. The original 25 images for the reference time period between 1984 and 2009 were taken from NASA's Landsat database. We fed these images into minimal pre-processing using a contrast normalization function to ensure equivalent image quality and then trained the SOM on the first image of the series. Image-by-image SOM produced consistently decreasing QE values, reliably reflecting the critical structural change around Lake Mead (Dresp-Langley, Wandeto, and Nyongesa, 2018 a and b). The lake is an artificial reservoir enclosed by the Hoover Dam. It collects water from the Colorado River, providing sustenance in water supply to the whole of Nevada and beyond. While the monotonous desert landscapes further away from the lake have not changed significantly in the years of the reference time period from which the image time series used here were taken, the immediate surrounds of Lake Mead became increasingly arid as the water levels of the lake progressively dwindled away. Linear trend analyses and correlation statistics (Pearson's product moment $P=0.957$; $p<.001$) showed a significant link between the QE, the natural phenomenon of drought reducing the surface covered by the lake year by year, and the water level statistics from the Hoover Dam Control Room for the same reference time period (Figure 3) .

The previous work had shown consistently that the SOM-QE proves a reliable indicator of the magnitude and the direction of local change in spatial contrast across images (Dresp-Langley, Wandeto, and Nyongesa, 2018 a). Some of the subtle variations in contrast contents across images were undetectable psychophysically in the light of signal detection theory (Green and Swets, 1966), even by expert radiologists, while the SOM-QE reliably detected and scaled these invisible changes (Wandeto, Nyongesa and Dresp-Langley, 2017; Wandeto, Nyongesa, Remond, and Dresp-Langley, 2017).

The new studies here explore the functional properties of the SOM-QE further and up to the single-pixel level in series of computer-generated achromatic and chromatic random-dot images. To further highlight the fine sensitivity of the unsupervised SOM-QE neural network metric to single-pixel  variations in images with several millions of pixels, we compared its performance to the RGB Mean, computed here with the most up-to-date image analysis tool currently available, *ImageJ*, which was developed and made freely accessible by the National Institute of Health (NIH). It will be shown that the RGB Mean performs in a similar way as the SOM-QE in signalling percent changes in RGB and single channel intensities across images when more than 10 out of 4 710 480 image pixels change, while the SOM-QE crushingly outperforms the RGB Mean when the task is to detect contrast change in a single pixel.

**Materials and methods**

Achromatic and chromatic random-dot image series and one random colour-image series, each with one original reference image (for training the SOM), and between five and 100 test images in a given series, were generated. After training the SOM on the original reference image, the other images from each series were fed into SOM to determine the QE as a function of systematic change in contrast contents across images. The first image of each of the different series will be referred to here as the "original" image (see Figures 4 and 5, for illustration) used for training the SOM. Several image series, with achromatic and chromatic contrast variations and other specific properties were analyzed.

*Achromatic random-dot images*
The original of these image series displays a randomly distributed number of white, black, light gray and dark gray dots on a medium gray background. The corresponding RGB values are given in Table 1. The total image area of these random dot images was 4 710 480 pixels. The diameter ($d$) of each single dot therein was 12 pixels. The single dot size (area) is obtained by

$$(d/2)^2 \times \pi \qquad\qquad\qquad (3)$$

where $\pi = 3.14$, which gives a single-dot size of 113 pixels in all the random-dot images. In the five test images of the first series (Figure 4 A1-A5), the area of a single black dot was increased systematically by 10% (Figure 4 A1), 20%, 30%, 40% and 60% (Figure 4 A5). In the second series (Figure 4 B1-B5), the area of a single white dot was increased systematically across the five images by the same magnitudes as here above. In the third series (Figure 4 C1-C5), the relative number of dark dots across images was increased by 10% (Figure 4 C1), 20%, 30%, 40% and 50% (Figure 4 C5) by increasing the number of black dots and holding the number of white and gray dots constant across the five images. In the fourth series (Figure 4 D1-D5), the relative number of white dots across images was increased by 10% (Figure 4 D1), 20%, 30%, 40% and 50% (Figure 4 D5) by increasing the number of white dots and holding the number of black and gray dots constant across the five images. In the fifth image series (Figure 5 A1-A5), a single black dot was shifted progressively towards the left across images by 20 pixels (Figure 5 A1), 40 pixels, 60 pixels, 80 pixels and 100 pixels (Figure 5 A5). In the sixth image series (Figure 5 B1-B5), a single white dot was shifted progressively towards the left across images by 20 pixels (Figure 5 B1), 40 pixels, 60 pixels, 80 pixels and 100 pixels (Figure 5 B5).

*Chromatic random-dot images*

In the chromatic image series, the random dots were given a specific colour (red, green and blue) of maximum intensity value (Table 1) within the specific single channel (R, G or B). This yields colour dot images of three different colours. At identical R, G, and B intensities (255), each of the three colours produces a markedly different perceived (subjective) intensity, or brightness, in each of the three reference images (Figure 6 A1, B1 and C1). In the three image series here (Figure 6 A1-A4, B1-B4 and C1-C4) the colour value of a single dot was progressively decreased in the corresponding channel, from the original value 255 in the reference image to 150, to 100 and to 50 across the other three images (2-4 respectively) of a given series.

*Chromatic single-pixel change*

To test for single-pixel change detection, all random-dots were set to gray background RGB (Table 1) and the G channel of a single pixel from the total of 4 710 480 image pixels was progressively varied between G=1 to 30, between G= 55-80, and between G=210-255 across a series of 100 images.

*Chromatic random noise with single-pixel removal*

This series of 70 images consisted of a random-colour pattern noise (Figure 7) with a total image area of 1 040 111 pixels. In the images of this series, single pixels were "removed" one-by-one (i.e. in the last image of this series of 70 a total of 69 pixels had been "removed") from arbitrary colours in the pattern by setting the single-pixel RGB to R=0, G=0, B=0.

*SOM training and analysis*

In each of the image series here, the training process consisted of 10 000 iterations. The SOM was always two-dimensional rectangular map of 4 by 4 nodes, hence capable of creating 16 models of observation from the data (cf. Figure 1). The spatial locations, or coordinates, of each of the 16 models or domains, placed at different locations on the map, exhibit characteristics that make each one different from all the others. When new input is fed into the map, the models compete and the winner will be the model whose features most closely resemble those of the input. An input will thus be classified or grouped by the models. Each model or domain acts like a separate decoder for the same input, i.e. independently interprets the information carried by a new input. The input is represented as a mathematical vector of the same format as that of the models in the map. Therefore, it is the presence or absence of an active response at a specific map location and not so much the exact input-output signal transformation or magnitude of the response that provides the interpretation of the input. To obtain the initial values for the map size, a trial-and-error process was implemented. It was found that map sizes larger than 4 by 4 produced observations where some models ended up empty, which meant that these models did not attract any input by the end of the training. It was therefore concluded that 16 models were sufficient to represent all the fine structures in the image data. The values of the neighborhood distance and the learning rate were set at 1.2 and 0.2 respectively. These values were obtained through the trial-and-error method after testing the quality of the first guess, which is directly determined by the value of the resulting quantization error ; the lower this value, the better the first guess. It is worthwhile pointing out that the models were initialized by randomly picking vectors from the training image, called the "original image" herein. This allows the SOM to work on the original data without any prior assumptions about a level of organization within the data. This, however, requires to start with a wider neighborhood function and a bigger learning-rate factor than in procedures where initial values for model vectors are pre-selected (Kohonen, 1998). The procedure described here is economical in terms of computation times, which constitutes one of its major advantages for rapid change/no change detection on the basis of even larger sets of image data before further human intervention or decision making. The times for training the SOM on the original images of the series from this study here range between 3.97 and 4.1 seconds. The times for subsequent SOM analysis of each test image from a series in view of generating the QE value range between 0.15 and 0.16 seconds. Thus, analysis of a series of 100 test images would take about 16 seconds after training the SOM (Wandeto, 2018). The code used for implementing the SOM is freely accessible at:

https://www.researchgate.net/publication/330500541_Self-organizing_map-based_quantization_error_from_images

*SOM-QE versus RGB Mean*

For RGB Mean calculations we used *ImageJ,* developed by the NIH *https://imagej.nih.gov/ij/* The average gray (RGB) level, or RGB Mean, in each single image from a given series corresponds to the sum of all gray values (RGB) of image pixels divided by the total number of image pixels. For RGB color images, the software converts each color pixel (*CP*) to grayscale beforehand, using

*CP =0.299 x* R *+ 0.587 x* G *+ 0.114 x* B          (4)

In order to compare how well the SOM-QE from performs in comparison with non-learning single-image analysis of the RGB Mean, all the images from the different series were uploaded one by one to *Image-J* and the mean for each image was computed.

*Data analysis*

The QE values from the SOM output and the image means from the *Image-J* analyses were stored in excel files and submitted to a comparative analysis as a function of the different systematic variations introduced in the test images of each of the nine image series.

**Results and Discussion**

The SOM-QE and the RGB Mean from analyses of the different image series were compared as function of the type of systematic variation in spatial image contrast contents.

*Achromatic random-dot series*

The SOM-QE and the RGB Mean distributions from the analyses of the mixed polarity random-dot image series were compared in the first instance. When analyzed as a function of a strictly local increase in the spatial extent (size) of a single dot in the achromatic image series shown in Figures 4 and 5, both the SOM-QE and the RGB mean are found to increase. This is highlighted further by the results from analyses as a function of a spatially distributed increase in relative contrast contents of either polarity across images. When the relative weight of opposite polarity contents changes in favour of positive polarity contents, by a systematically increasing number of white dots across images, the SOM-QE and the RGB mean are found to also increase systematically. Conversely, when the relative weight of opposite polarity contents changes in favour of negative polarity contents, by a systematically increasing number of black dots across images, the SOM-QE and the RGB mean are found to decrease systematically. These results are displayed in Figure 8. It is shown that both the SOM-QE and the RGB mean are consistently sensitive to the polarity of changes in the relative weight of contrast contents across images of a series. When analyzed as a function of changes in the spatial position of strictly local contrast contents in terms of shifts in the spatial position of a contrast dot across images in a series, both the SOM-QE and the RGB mean are found to be invariant, i.e. they do not capture changes in the local spatial position of image contrast contents of either sign. These results are displayed at the bottom of Figure 8. Both the SOM-QE and the RGB mean are insensitive to the spatial position of contrast detail in images. This finding has implications for image analysis. On the one hand it represents a limitation of the change detection potential of both metrics, as neither the SOM-QE nor the RGB mean signals change when the criterion is the spatial position of a specific contrast intensity of any sign, or a specific colour. On the other hand, in some contexts insensitivity to changes in the spatial position of a contrast or colour could represent an advantage. Radiological images from different time series, for example, are never taken from precisely the same spatial position. A qualified change detector in this respect would have to be able to discard image variability due to shifts in spatial position of specific contrasts or colour. The SOM-QE and the RGB mean both qualify in this respect. To assess the statistical significance of SOM-QE and RGB Mean variations, i.e. whether contrast or colour changes signalled by either metric differ significantly from some hypothetical ground state, one-sample t-tests were run on the data displayed in Figures 8 and 9. In a time series where potentially meaningful changes may occur from one moment (image) in time to another, the image reflecting the hypothetical ground state would be the first image from that series, or some image taken much earlier in time. Here, the ground state image, called the "original" image on which the SOM was trained, is always the first image of a given series. This first image reflects a ground state of "no change" before systematic alterations to the other images of a series were introduced. The one-sample t-tests run on the SOM-QE and RGB Mean distributions take the original (first) image of a given series as the hypothetical ground state or hypothetical population mean set as criterion for the statistical null hypothesis. The t-statistic will not be significant if the SOM-QE or the RGB Mean from the test images of a given image series do not vary significantly in comparison with the value from analysis of a given ground state image. The results from these statistical tests are summarized here Table 2. They consistently show significant $t$ statistics for both the SOM-QE distributions and the RGB Mean distributions from the different analyses plotted in Figures 8 and 9. As could be expected, the $t$ statistics relative to distributions for the achromatic image series with systematic gradual shifts in spatial position of contrasts are not significant.

*Chromatic random dot series*

The results from SOM-QE and RGB Mean analysis as a function of single channel colour (Red, Green, Blue) decrements in a random dot from the three chromatic image series (Figure 6) are displayed in the graphs in Figure 9. Both the SOM-QE and the RGB Mean are shown to capture the single channel decrements consistently for each of the three colours.

*Chromatic single-pixel change*

The SOM-QE and the RGB Mean, plotted as a function of single-pixel G channel variations across 100 gray images of a total image area of 4 710 480 pixels each, are displayed in the graphs in Figure 10. These results show that the SOM-QE consistently detects and scales the single-pixel G channel variations while the RGB Mean is insensitive to changes at the single-pixel level.

*Chromatic random noise with single-pixel removal*

The limitation in sensitivity of the RGB Mean to contrast changes at a fine spatial scale is highlighted further by the graphs in Figure 11, which compare between the SOM-QE and the RGB Mean as a function of progressive one-by-one single-pixel "removal" across 70 random-color noise images with a total area of 1 040 111 pixels (illustrated in Figure 7). Pixels from arbitrary colours in these image were progressively "removed" by setting the single-pixel RGB to R=0, G=0, B=0. The SOM-QE, as shown in the graphs in Figure 11, consistently detects each and every single-pixel "removal", the RGB Mean is insensitive to any change smaller than 10/1 040 111 pixels.

*Functional implications*

The ability of the SOM-QE to signal the finest changes in spatial contrast or colour across images with a to-the-single-pixel precision has potentially important implications in the wider context of biologically inspired neural networks for image analysis. The ability to discriminate with fine spatial precision between different levels of local contrast intensities and contrast signs is akin to known functional properties of a specific class of retinal ganglion cells (the so-called Y-cells) in the visual systems of the primate and the cat (i.e. Shapley and Perry, 1986). It is therefore legitimate to affirm that SOM-QE mimics some of the early stages of visual information processing in the primate brain (Figure 12) and, in addition, is capable of performing beyond the capacity limits of human visual detection. This conveys to the SOM-QE an hitherto unsuspected and potentially important functional property.

The visual analysis of changes in spatial position, on the other hand, requires the higher-order functional properties of orientation-selective visual cortical neurons (Hubel and Wiesel, 1959, 1965, 1968), only some of which are selective to contrast polarity (the so-called simple cells) while others (the so-called complex cells) are not (e.g. Zhang & von der Heydt, 2010; Su et al., 2010). Long-range interactions between cortical neurons (Spillman, Dresp-Langley, and Tseng, 2015, for a comprehensive review) may allow for a finely tuned visual integration of even the smallest local changes in spatial position. This potential of cortical visual neurons is reflected by a well-studied human ability called "visual hyperacuity", which is the ability to detect changes in the local spatial position of contrasts ten times smaller than the narrowest spacing of visual receptors in the retina (Westheimer, 1981). This functional property of human vision is partly a result of perceptual learning, and requires modeling by deep neural network structures with complex functional architectures (e.g. Wenliang & Seitz, 2018) akin to those developed already much earlier by Steve Grossberg and colleagues (e.g. Grossberg, 2013, 2015, for reviews).

**Conclusions**

The quantization error in Self-Organizing Maps with a relatively simple functional architecture can be effectively exploited as an indicator for the rapid automatic detection of systematic and potentially significant changes in images from large time series, albeit with clear limitations. The indicator will reliably signal change in situations where potentially critical information is removed from or added to a scene, but will not signal change when relative and total amounts of information are constant, but the spatial location of a contrast element in the scene is changed. This may represent either an advantage or a disadvantage for automatic change detection, depending on what is required. In situations where small shifts in spatial location of image contents are irrelevant, the indicator qualifies as an effective quantitative measure. In summary, the quantization error in Self-Organizing Maps displays reliable sensitivity to the relative spatial extent, intensity, and polarity of local contrast in images akin to that of visual neurons (Y-cells) in the retina. This level of visual analysis does not take into account the higher-order cortical neural mechanisms exploited in complex neural network architectures for human vision, like those developed by Steve Grossberg and his colleagues. The QE in SOM does not seem to capture detailed shape information or spatial location and is therefore not directly exploitable whenever these latter need to be taken into account. Finally, from a user perspective and that of potential applications, it may be worthwhile pointing out that the SOM-QE approach as exploited here in this work does not require human intervention for image annotation, classification, selection of regions of interest in the image, or change criterion setting. The approach is, in this respect, fully automatic and therefore objective. The SOM-QE approach described here takes a whole series of images as it is, without further human intervention, and automatically sets criteria for change or no change in terms of the QE associated with the final synaptic weights and after unsupervised learning. These criteria then become the benchmarks against which each subsequent image is compared. The consistently fine sensitivity of the SOM-QE to single-pixel-change in long series of large images has potential within the context of unsupervised image learning and computational building block approaches for fully automatized image analysis. This includes deep learning blocks for large unsorted image data sets, and the automatic detection of contrast changes at the nanoscale in Transmission or Scanning Electron Micrographs (e.g. Wandeto & Dresp-Langley, 2019), or at the subpixel level in multispectral and hyper-spectral imaging data (e.g. Kerekes & Baum, 2002).

**Acknowledgments**

We would like to thank the Guest Editor Donald Wunsch and two anonymous reviewers for their helpful comments and suggestions. Our Norwegian colleagues O. V. Solberg and O. K. Ekseth provided further advice and encouragement, for which we are most grateful.

**Tables and Figures with captions**

|  | *R* | *G* | *B* |
|---|---|---|---|
| *White* | 255 | 255 | 255 |
| *Black* | 0 | 0 | 0 |
| *Dark gray* | 127 | 127 | 127 |
| *Light gray* | 191 | 191 | 191 |
| *Red dots* | 255 | 0 | 0 |
| *Green* | 0 | 255 | 0 |
| *Blue* | 0 | 0 | 255 |
| *Gray background pixels* | 179 | 179 | 179 |

Table 1

RGB values of white, black, light gray and dark gray pixels and of the medium gray image background pixels.

| ***SOM-QE*** | *dot size+ sign +* | *dot size+ sign -* | *n dots+ sign+* | *ndots+ sign -* | *pos shift sign+* | *pos shift sign -* |
|---|---|---|---|---|---|---|
| *N images* | 6 | 6 | 6 | 6 | 6 | 6 |
| *Normality Test* | .965 | .963 | .949 | .940 | .056 | .064 |
| *DF* | 5 | 5 | 5 | 5 | 5 | 5 |
| *One-sample 't'* | 3.264 | 3.277 | 3.025 | 2.998 | 1.973 | 2.125 |
| *Probability* | $p < .05$ | $p <.05$ | $p <.05$ | $p <.05$ | NS | NS |
| *95 % confidence* | .392-3.302 | .403-3.340 | .081-.992 | .0777-1.013 | .001-.007 | .0007-.0008 |

| ***RGB Mean*** | *dot size+ sign +* | *dot size+ sign -* | *n dots+ sign+* | *ndots+ sign -* | *pos shift sign+* | *pos shift sign -* |
|---|---|---|---|---|---|---|
| *N images* | 6 | 6 | 6 | 6 | 6 | 6 |
| *Normality Test* | .835 | .815 | .963 | .956 | Failed | Failed |
| *DF* | 5 | 5 | 5 | 5 | 5 | 5 |
| *One-sample 't'* | 2.935 | 2.863 | 3.287 | 3.265 | -- | -- |
| *Probability* | $p < .05$ | $p <.05$ | $p <.05$ | $p <.05$ | -- | -- |
| *95 % confidence* | .00120-.0181 | .00226-.0421 | .0119-.100 | .0122-.100 | -- | -- |

Table 2

Results of the one-sample *t*-tests on the SOM-QE (top) and the RGB Mean (bottom) distributions for images (*N*=6, including the original) from the six first series.

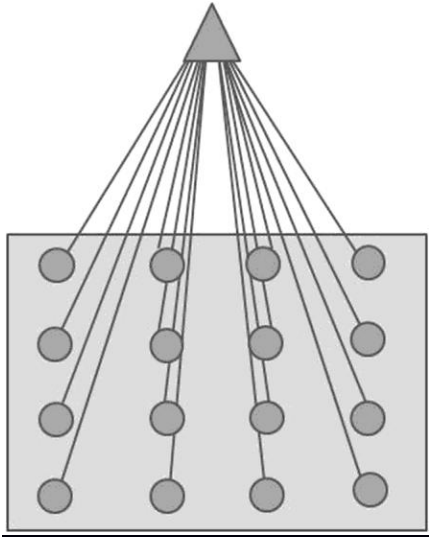

Figure 1: Representation of a SOM prototype with 16 models, indicated by the filled circles in the grey box. Each of these models is compared to the SOM input in the training (unsupervised winner-take-all learning) process. The input vector corresponds to the RGB image pixel space; the model in the map best matching the SOM input will be a winner, and the parameters of the winning model will change to further approach the input. Parameters of models within close neighborhood of the winning model will also change, but to a lesser extent compared with those of the winner. At the end of the training, each input space will be associated with a model within the map. The difference between input vector and the winning model determines the quantization error in the SOM output.

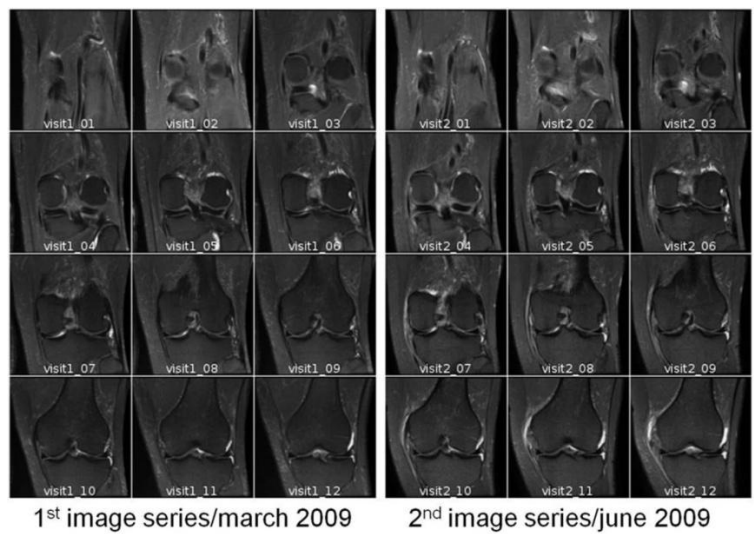

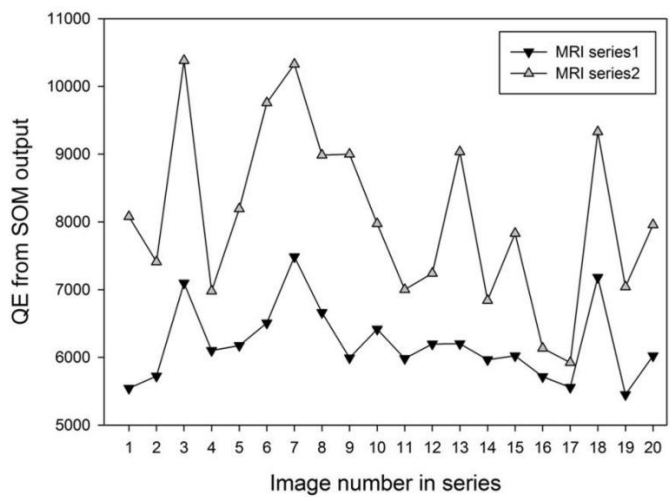

Figure 2

After training a SOM on the first image of a first MRI time series (20 images per series; only 12 from each clinical visit are displayed here for ilustration) SOM analysis was performed on the other images in the series, taken from two clinical examinations of a patient's kee, before and after a minor accident. The QE from these SOM analyses reliably reflects systematic changes in image contrast contents between the two time series, related to local tissue damage caused by the accident (from Wandeto, Nyongesa, Remond, and Dresp-Langley, 2017). These changes were not immediately detected by human visual inspection.

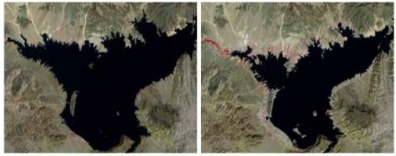
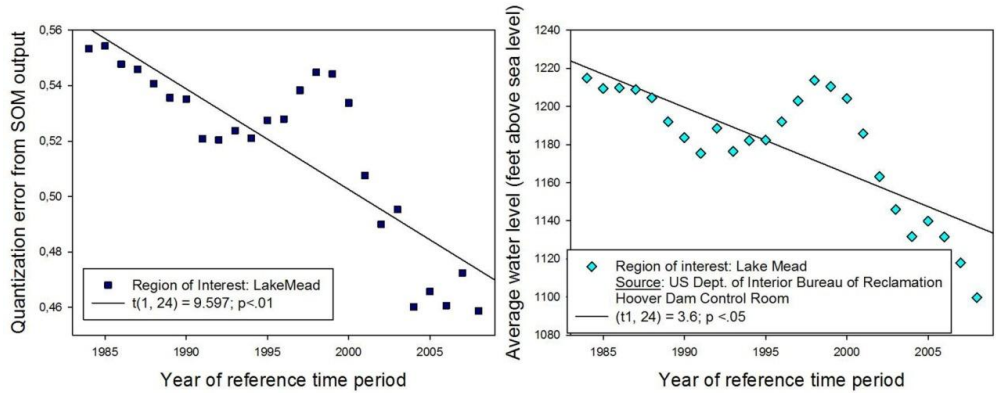

Figure 3

Results from analyses of times series of satellite images of Lake Mead in the Nevada desert, taken across the years 1984-2009. The two example on top show the first and the last images from that series. The QE from the SOM analyses is shown as a function of the year in which a given image was taken (graph on the left), and as a function of the corresponding water level statistics from the Hoover Dam Control Room for the reference time period (graph on the right). Linear trend analysis and a one-sample t-statistic signal a statistically significant decrease in the QE from the image analyses, and in the independently measured water levels. Correlation statistics (Pearson's product moment P=0.957; p<.001) showed a significant link between the QE, reflecting the natural phenomenon of drought reducing the surface covered by the lake year by year, and the water level statistics (from Dresp-Langley, Wandeto, and Nyongesa, 2018 a and b).

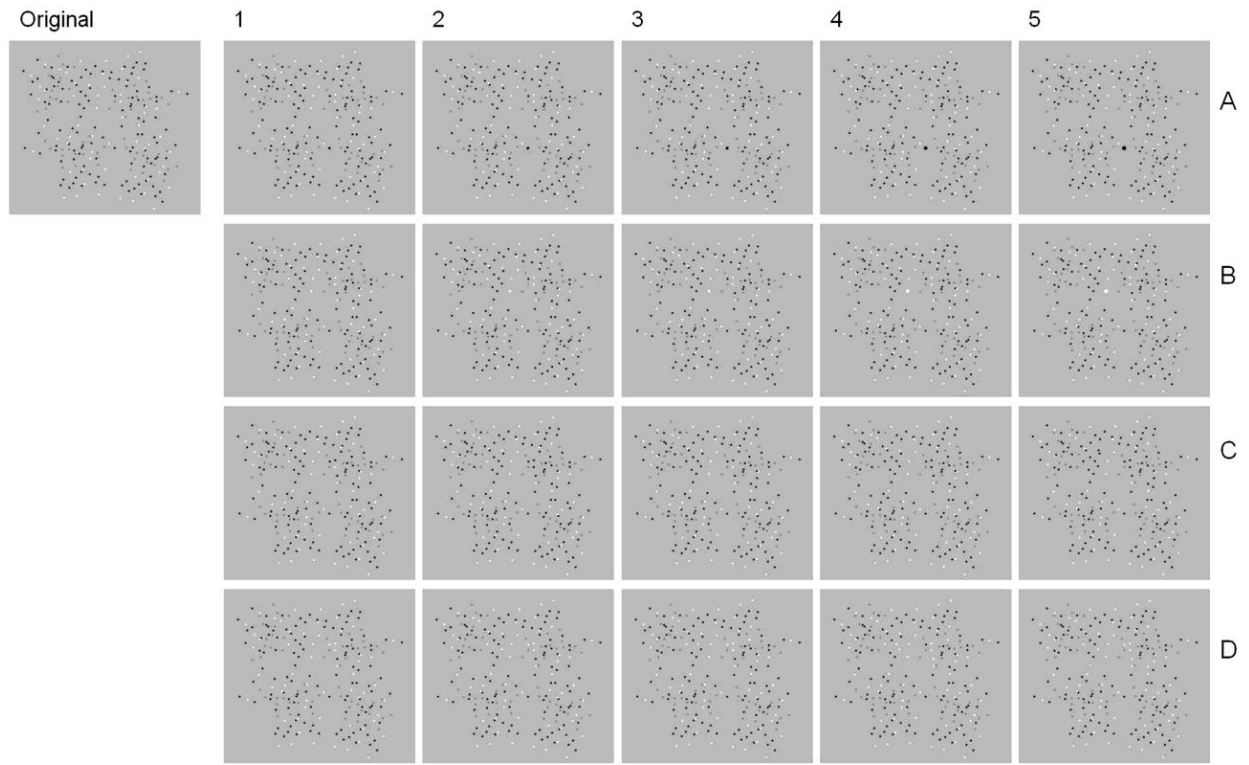

Figure 4

Four random-dot image series from this study. One original reference image (for training the SOM) and five test images in each series were computer generated. After training the SOM on the original reference image, the five others from each series were submitted to SOM. Variations in QE were studied as a function of systematic local (single dot) variations in spatial contrast contents, some of them visually undetectable, across the images of a given series.

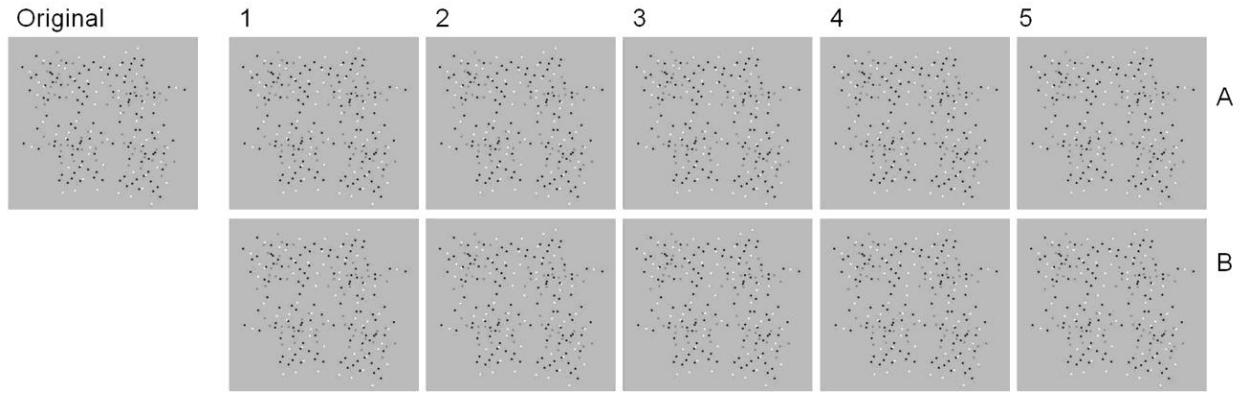

Figure 5

Two random-dot image series from this study. In the five test images of each series here, the local spatial position of a single dot with negative contrast polarity (A) or with positive contrast polarity (B) was systematically varied.

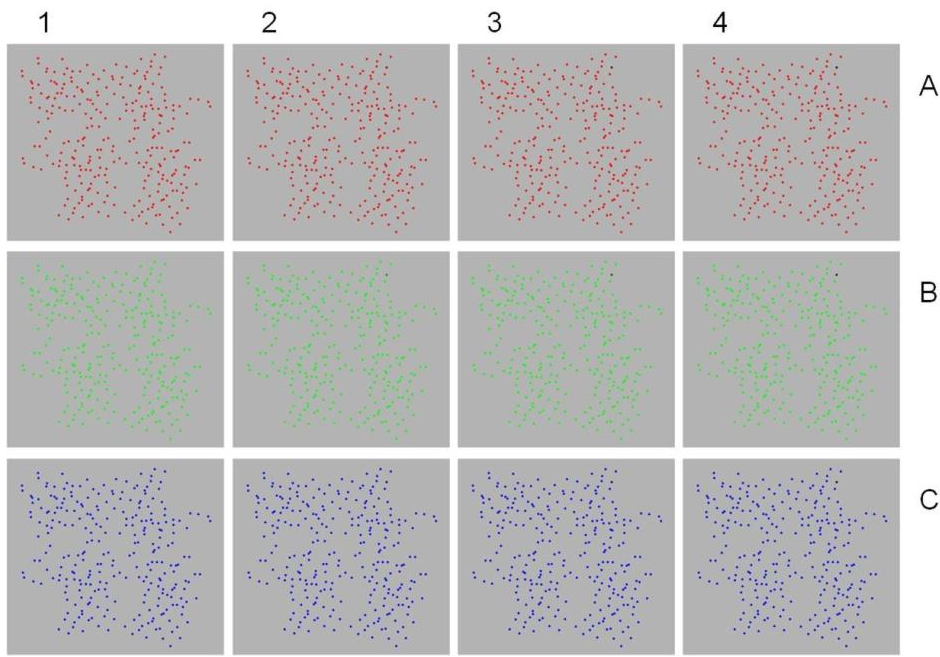

Figure 6

Three random-dot image series (A, B, C) with single channel colours (Red, Green, Blue) set at the maximum level (R=255 in image A1; G=255 in image B1; B=255 in image C1) in the first of the four images of a given series. In images 2, 3 and 4, the channel colour value of a single dot is decremented to 150, 100, and 50. The colour decrement is easily perceived visually in the case of the green dots (image series B1-4), to a lesser extent in the case of the red dots (image series A1-4) and hardly if at all in the case of the blue dots (image series C1-4).

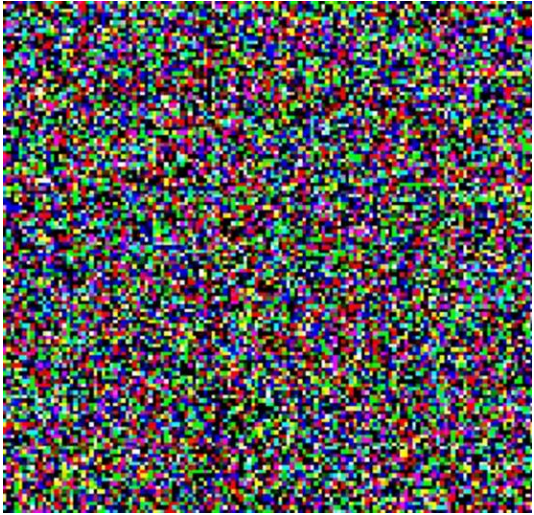

Figure 7

Original image of the last series of 70 images from the experiments here. Here, the total image area consists of 1 040 111 pixels. In the 69 subsequent images of this series, single pixels were "removed" one-by-one (i.e. in the last image of this series a total of 69 out of the 1 040 111 pixels had been "removed"). Pixels from arbitrary colours in the image were "removed" by setting the single-pixel RGB to R=0, G=0, B=0.

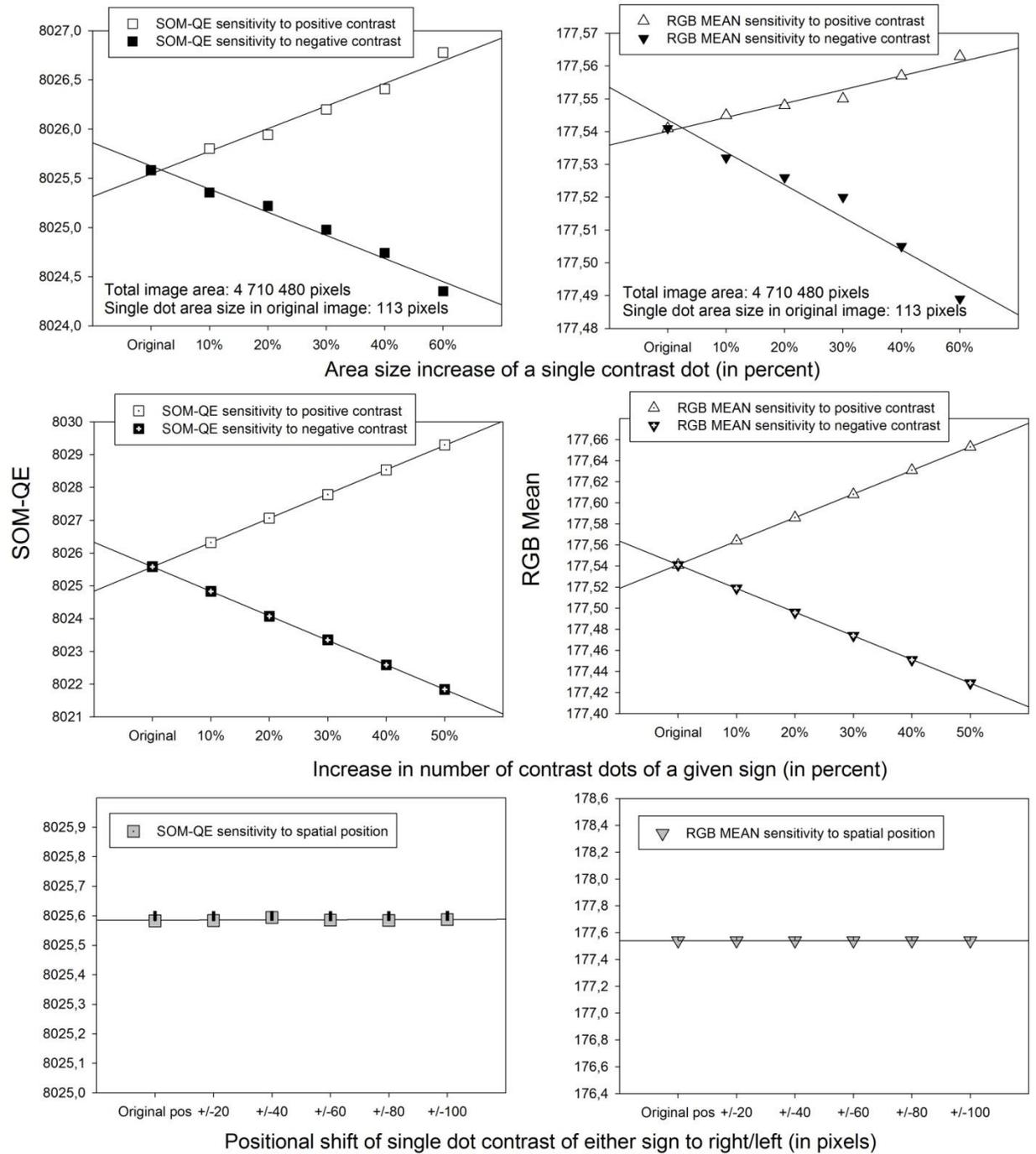

Figure 8

The SOM-QE (left) and the RGB Mean (right) as a function of the percent increase in a) single contrast dot area (113 pixels in original training image; total image area: 4 710 480 pixels) towards positive and negative polarities (top), b) relative amount of either polarity reflected by the relative number of white and black dots (middle), and c) spatial position shift across images towards left or right of a single contrast dot with positive or negative contrast polarity (bottom).

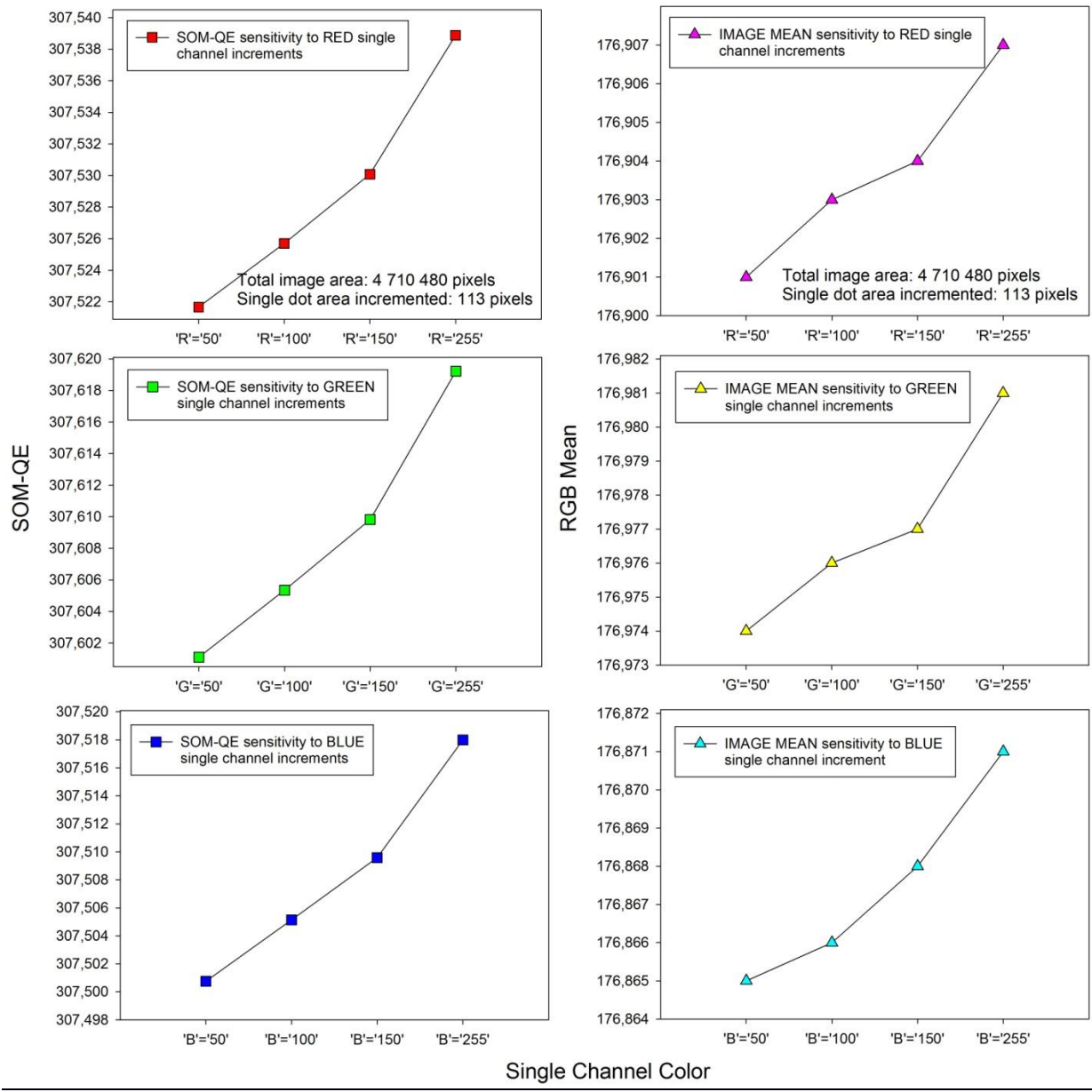

Figure 9

The SOM-QE (left) and the RGB Mean (right) as a function of the single channel colour (Red, Green, or Blue) of one random-dot of constant size (dot area = 113 pixels) in the image. The single channel (R, G or B) of a random-dot in the image is progressively decremented across images. Both SOM- QE and the RGB Mean are sensitive to these single channel decrements.

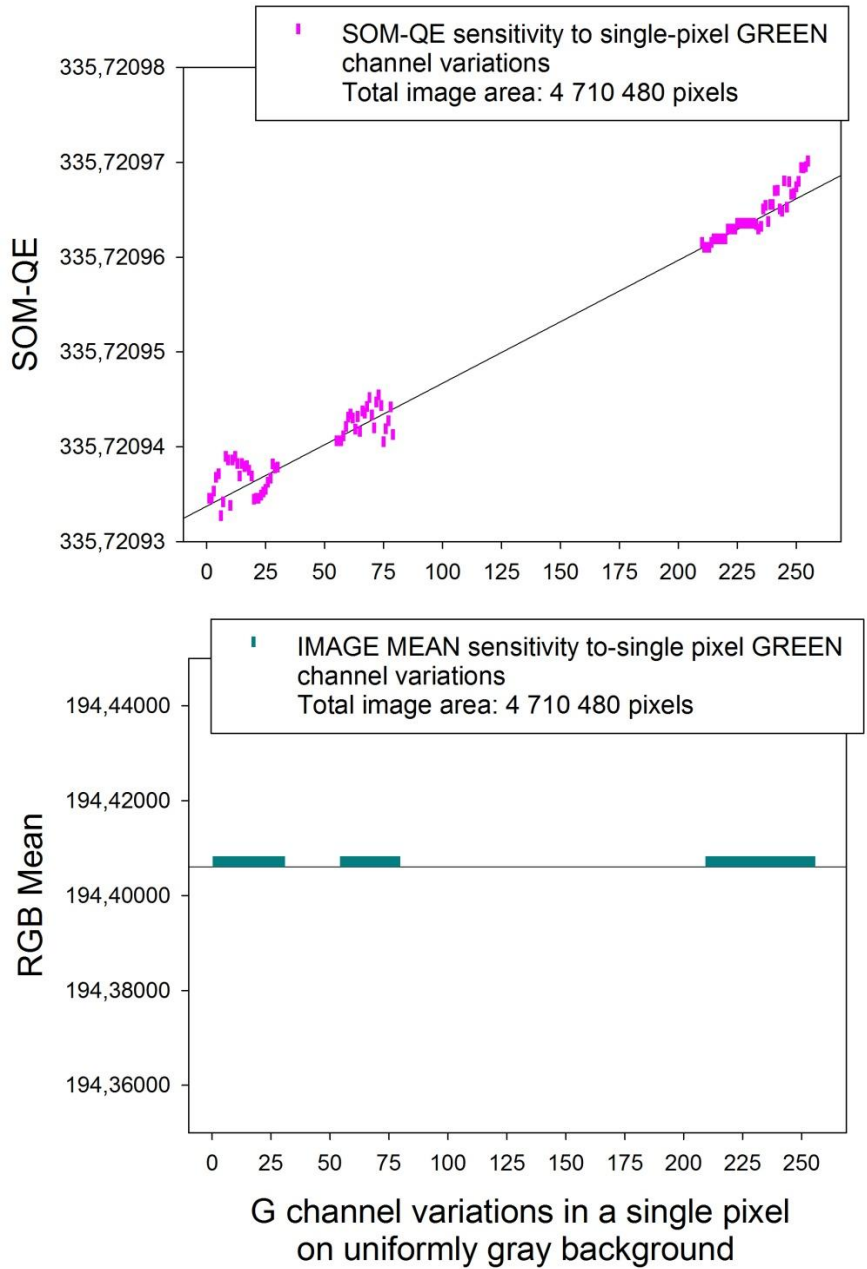

Figure 10

The SOM-QE (left) and the RGB Mean (right) as a function of single-pixel G channel variations across 100 gray images of a total image area of 4 710 480 pixels each. The 4 710 479 non varying gray pixels are set at R=179, G=179, B=179). The SOM-QE consistently detects and scales the single-pixel G channel variations, the RGB Mean is insensitive to such changes at the single-pixel level.

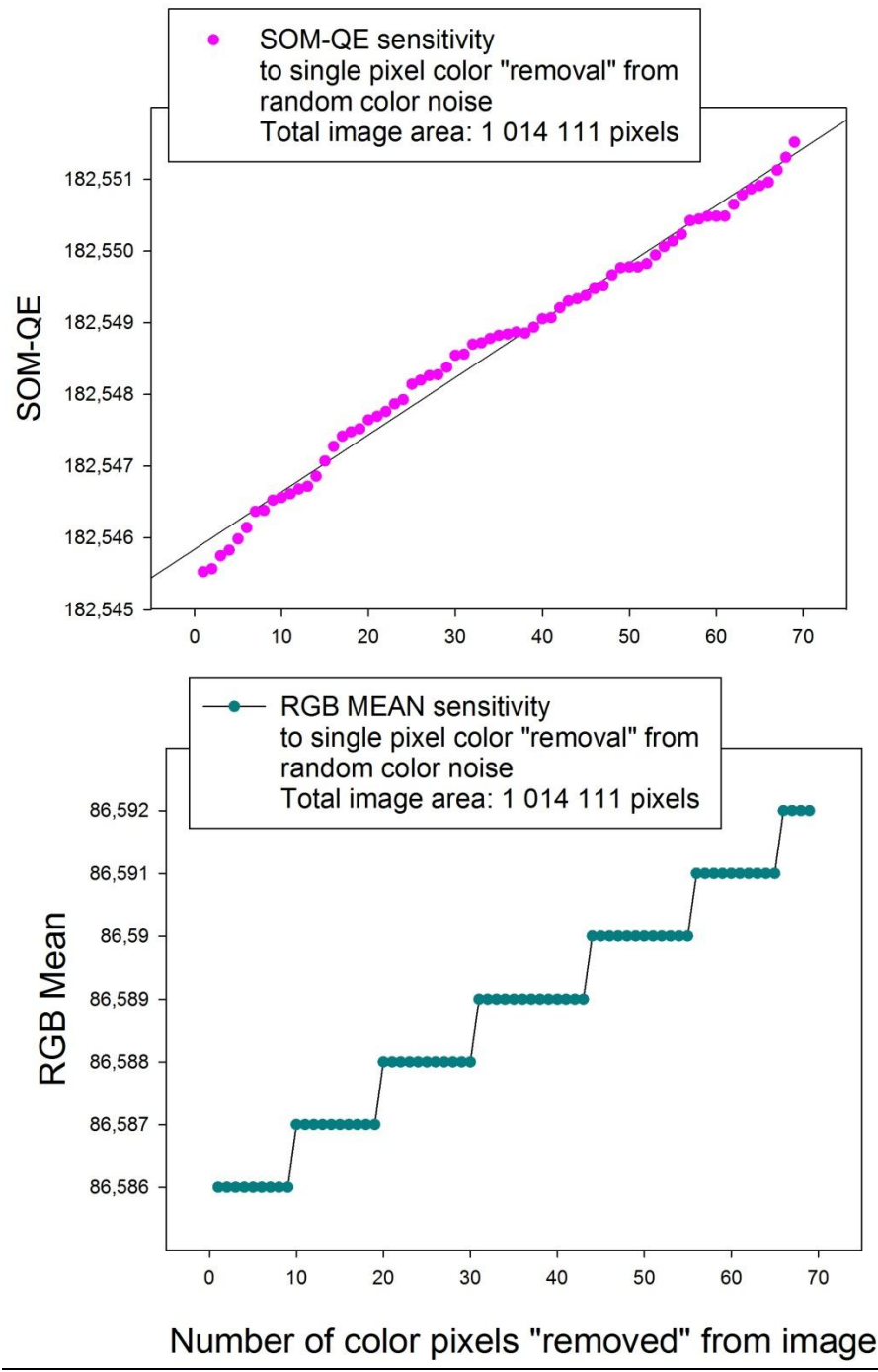

Figure 11

The SOM-QE (left) and the RGB Mean (right) as a function of progressive one-by-one single-pixel "removal" across 70 random-color (Fig. 6) images with a total area of 1 040 111 pixels. Pixels from arbitrary colours in the image were progressively "removed" by setting the single-pixel RGB to R=0, G=0, B=0. The SOM-QE consistently detects single-pixel "removal", the RGB Mean is insensitive to changes smaller than 10 pixels.

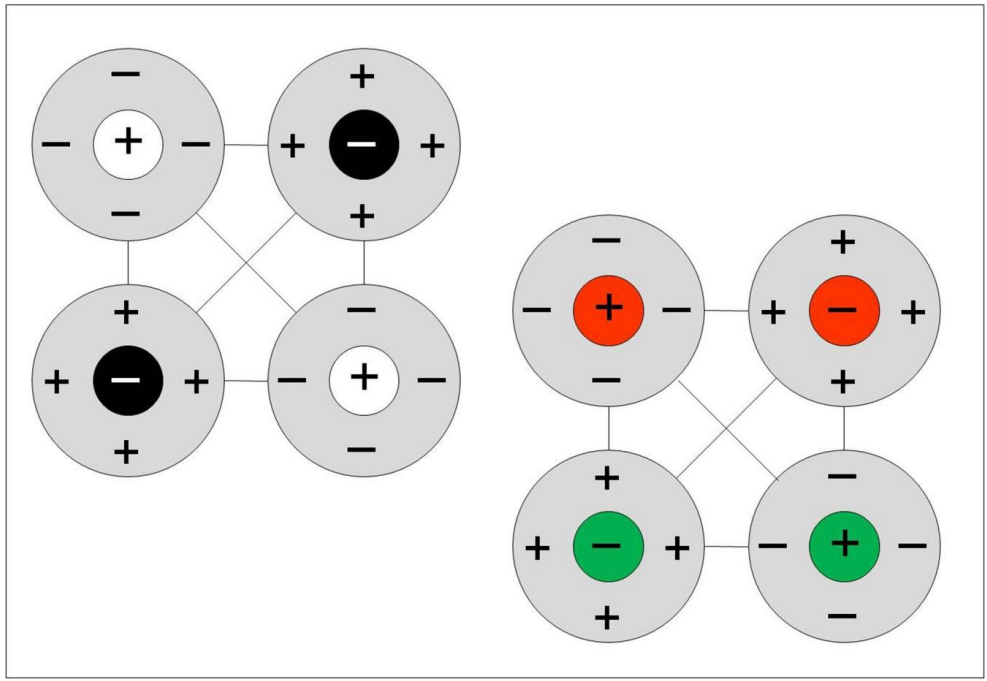

Figure 12

The early processing of contrast intensity and contrast sign by the visual system of primates and the cat is ensured by fully connected neural networks of retinal ganglion cells of the Y-type (cf. Shapley & Perry, 1986), with functionally identified antagonistic centre-surround receptive field organization. In the illustration on the left, a network response of the '*ON*' type to contrasts of positive sign on darker backgrounds is represented by '+', a network response of the '*OFF*' type to contrasts of negative sign on lighter backgrounds is represented by '-'. Colour contrasts may be processed by similar antagonistic mechanisms at this level, as suggested by the illustration on the right for the case of RED and GREEN. The sensitivity of the SOM-QE to changes in the local intensity and sign of spatial contrast in images mimics the functional properties of such antagonistic neural networks.